\RequirePackage{amsthm}
\documentclass[sn-mathphys-num]{sn-jnl}

\usepackage{graphicx}%
\usepackage{multirow}%
\usepackage{amsmath,amssymb,amsfonts}%
\usepackage{amsthm}%
\usepackage{mathrsfs}%
\usepackage[title]{appendix}%
\usepackage{xcolor}%
\usepackage{textcomp}%
\usepackage{manyfoot}%
\usepackage{booktabs}%
\usepackage{algorithm}%
\usepackage{algorithmicx}%
\usepackage{algpseudocode}%
\usepackage{listings}%
\usepackage{svg}%

\usepackage[binary-units]{siunitx}

\theoremstyle{thmstyleone}%

\theoremstyle{thmstyletwo}%

\theoremstyle{thmstylethree}%

\begin{document}

\title[Resource efficient data transmission on animals based on machine learning]{Resource efficient data transmission on animals based on machine learning}

\author*[1,2]{\fnm{Wilhelm} \sur{Kerle-Malcharek}}\email{wilhelm.kerle@uni-konstanz.de}

\author[1]{\fnm{Karsten} \sur{Klein}}\email{karsten.klein@uni-konstanz.de}

\author[2,4]{\fnm{Martin} \sur{Wikelski}}\email{wikelski@ab.mpg.de}

\author[1,3]{\fnm{Falk} \sur{Schreiber}}\email{falk.schreiber@uni-konstanz.de}

\author[2]{\fnm{Timm A.} \sur{Wild}}\email{twild@ab.mpg.de}

\affil*[1]{\orgdiv{Life Science Informatics}, \orgname{University of Konstanz}, \orgaddress{\city{Konstanz}, \postcode{78464}, \country{Germany}}}

\affil[2]{\orgdiv{Wild Lab}, \orgname{Department of Migration, Max Planck Institute of Animal Behavior}, 
\orgaddress{\city{Radolfzell}, \postcode{78315}, \country{Germany}}}

\affil[3]{\orgdiv{Faculty of Information Technology}, \orgname{Monash University}, \orgaddress{\city{Clayton}, \postcode{3000}, \country{Australia}}}

\affil*[4]{\orgdiv{Department of Biology}, \orgname{University of Konstanz}, \orgaddress{\city{Konstanz}, \postcode{78464}, \country{Germany}}}

\abstract{\textbf{Background:} 
  Bio-loggers, electronic devices used to track animal behaviour through various sensors, have become essential in wildlife research.
  Despite continuous improvements in their capabilities, bio-loggers still face significant limitations in storage, processing, and data transmission due to the constraints of size and weight, which are necessary to avoid disturbing the animals.
  This study aims to explore how selective data transmission, guided by machine learning, can reduce the energy consumption of bio-loggers, thereby extending their operational lifespan without requiring hardware modifications.
    
\textbf{Methods:} 
 Our study employs machine learning techniques, specifically decision trees, to recognize a chosen animal behaviour from sensor readings.
 We collected data from human behaviour as a less complex example to create the theoretical foundation of our approach.
 These decision trees are trained to classify behaviours and determine the most relevant data to transmit based on the classification.
 Various decision tree models, with different combinations of sensors as input values, were produced and tested to investigate the impact of different sensors on the classification.
 The models were evaluated based on their ability to maintain high classification accuracy while reducing the overall computation cost, if possible.
 Lastly, we evaluate the reduction of energy consumption, based on a state-of-the-art bio-logger, the WildFi tag.  
 
\textbf{Results:} 
  We demonstrate that decision trees can be an effective tool for enabling bio-loggers to detect specific behaviours with accuracies above $80\%$ autonomously and selectively transmit only essential data.
  Our study also reveals that a subset of sensor features can result in minor precision reductions of less than $2\%$, but major reductions of $20\%$ of data in required input, recorded and transmitted data.
  Furthermore, we illustrate the substantially higher cost of data transmission compared to filtering the data beforehand, which is 10 times cheaper for the WildFi tag even for tiny data packages.
  
\textbf{Conclusion:}    
  Our study underscores the potential of machine learning to optimize the energy consumption of bio-loggers by controlling data transmission.
  The approach offers a promising pathway for enhancing the longevity of bio-loggers, facilitating longer-term animal behaviour studies and ultimately contributing to more sustainable and efficient wildlife monitoring practices.

}


\maketitle

\section{Background}\label{sec1}
The study of animal behaviour has a rich history, driven by the interest in understanding the mechanisms and reasons behind an animals' decision making~\cite{richter1927animal,beach1954effects,mcfarland1971feedback,manning2012introduction}.
Advances in technology have significantly contributed to corresponding research, allowing for detailed data collection.
For instance, measuring the diving capacities of Weddell seals~\cite{Kooyman1965}, GPS tracking of birds and wolves~\cite{GUILFORD2008,kojola2006dispersal,sand2005using}, and tracking the acceleration patterns of fish and other animals over time~\cite{brown2013observing,GLEISS201085,kroschel2017remote,Krone2008} to infer knowledge about their behaviour ultimately are possible nowadays.
The electronic recording devices which enable such recordings and are attached to animals are called bio-loggers~\cite{hooker2007bio}.
Bio-loggers tremendously increased the understanding of animal behaviour, fueling the interest in further advancing bio-logging technologies.

In particular, inertial measurement units (IMUs), which include 3-axis accelerometers, magnetometers, and or gyroscopes, have become essential in studying animal movements~\cite{andrzejaczek2019biologging,Fourati2011,ware2016averaged,chakravarty2019novel,chakravarty2020seek}.
State-of-the-art bio-loggers, such as the LoRaWAN bio-logger~\cite{Gauld2023} and the WildFi tag~\cite{Wild2022}, integrate such IMUs with GPS and other sensors to gather information-rich datasets.
Naturally, more sensors that capture information with increasing resolution (more data points) also put a higher strain on the battery lifetime (more measurements require more energy) of bio-loggers, as well as their storage capacities (more readings produce larger file sizes).
Since the recorded information becomes denser, novel data transmission approaches gained popularity and advancements to circumvent the higher storage demands by simply freeing up storage of data that has already been transmitted~\cite{ayele2018towards,Wild2022,krondorf2022icarus,Bridge2011,Kays2015,wild2023multi}.
The downside to data transmission is that it is one of the most energy-consuming activities of distributed embedded systems~\cite{AZAR2019168}.
Thus, while WiFi technology like it is used by the WildFi tag allows for a transmission speed of about $230\si{\kilo\byte}/s$, the bio-logger runtime shortens compared to just storing data.
For comparison: A WildFi tag measuring with a 9-axes IMU and transmitting its data via WiFi has a reduced runtime of approximately 6 hours~\cite{Wild2022}.
Still, using WiFi protocols is an advancement essentially solving storage limitations, allowing retrieval of data without the need to re-capture the animal, and gathering fine-grained information, leveraging the systems to be capable of handling big data.
These challenges and possibilities necessitate a trade-off between data resolution and the device operating time, especially since bio-loggers must be small and unobtrusive to minimize disruption to the animals, which is why they typically have limited energy storage, constraining the data resolution or shortening operating time even further.
The Internet of Things (IoT), which concerns itself with networks of interconnected devices that communicate and exchange data, offers valuable strategies for managing energy consumption for those interconnected devices through software and hardware solutions~\cite{humayun2022energy,diene2020data,Saqlain2019}.
By employing sensor fusion, IoT systems can deduce specific conditions of tracked objects~\cite{salah2020iot} and optimize sensor usage and power management accordingly~\cite{humayun2022energy}.
This concept of smart energy management is particularly crucial for bio-loggers, as it offers ways to tackle the trade-off between data resolution and energy consumption.

Recent publications suggest that machine learning can be used to detect specific states of animals~\cite{Bidder2020}.
Generally, machine learning on animal behaviour time series data to classify their behaviour at a certain point in time has seen several approaches and is considered to hold a lot of potential~\cite{brandes2021behaviour,tuia2022perspectives}.
Bidder et al. used the k-nearest-neighbour algorithm to provide easy-to-use solutions for non-specialists in machine learning~\cite{bidder2014love}.
Chakravarty et al. present several approaches towards optimisations for behaviour recognition using accelerometers, like the use of high-fidelity data~\cite{chakravarty2020seek} or even looking at it at a biochemical level~\cite{chakravarty2019novel}.
As the topic unfolds for bio-loggers, different methodologies are being discussed and applied for them, as well~\cite{kadar2020assessment,wang2019machine}.
Given the potential of recognising a specific behaviour of an animal, researchers like Korpela et al. even transferred parts of the machine learning approaches onto the bio-logger and utilised it.
They explored the idea of on-board classification for preserving energy by limiting the recording of data to behaviours of interest.
They trained decision trees with the criterion that they exhibit a good balance of cost to accuracy and based their data on 3-axis accelerometers~\cite{Korpela2020}.
With our study, we want to further this thought.
We exemplarily show that on-board pattern recognition works and is facilitated through sensor fusion by integrating other sensors than accelerometers, like gyroscopes, which face neglect for the most part as researchers focus on established hardware.
We put into perspective the net costs of on-board classification to show, that the on-board recognition is feasible, too.
We offer comprehensive insights into how machine learning can be used to improve not only data sampling but also data transmission strategies for state-of-the-art bio-loggers.
We argue that the overall gain in runtime of the bio-logger can be more than doubled.

\section{Methods}\label{sec11}
    The general methodology we follow to reduce the energy consumption of bio-loggers is based on the premise that we want to reduce the time spent and the length of messages transmitted.
    To achieve this, we make use of pattern recognition with machine learning.
    We train a model to distinguish different animal behaviours based on sensor data, try to reduce the required data for recognition, and supply the bio-logger with the finished model.
    With the help of the classifier, the device makes decisions based on the recognised behaviours about which data to store and transmit, thus filtering the transmitted data.
    
    \subsection{Hardware}
        
\begin{figure}[h!]
        \centering
        \includegraphics[width=.7\textwidth]{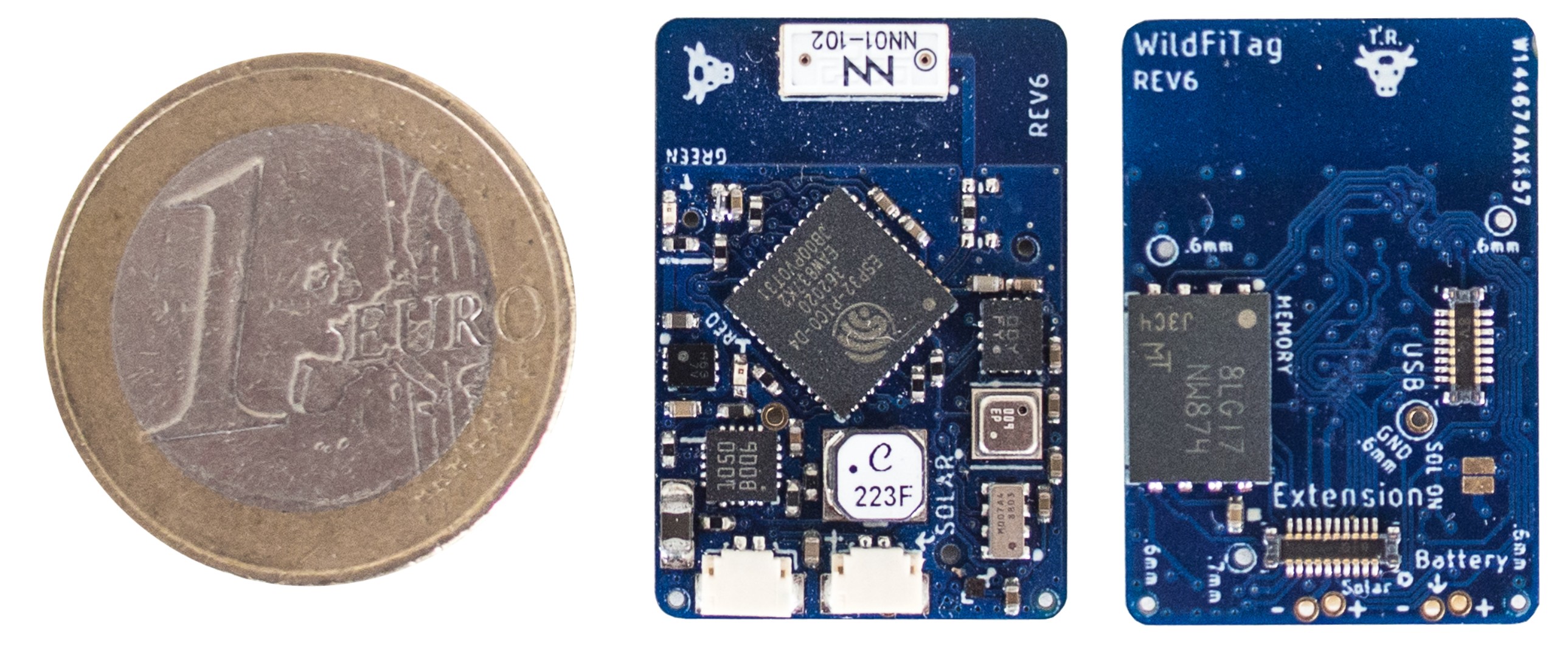}
        \caption{An illustration of the WildFi tag from the original publication~\cite{Wild2022}. On the left is a coin to give a perspective on the device's size. In the centre is the main device from front and, on the right, from back. }
        \label{fig:wildfiEx}
    \end{figure}

    The bio-logger we use is the WildFi tag~\cite{Wild2022} (Fig.~\ref{fig:wildfiEx}).
    The WildFi tag is a cutting-edge bio-logger utilising advances in the fields of IoT and bio-logging alike.
    The size of the logger is $25.95mm\times17.85mm\times0.6mm$, and its weight is about 1.28g without the GPS extension, which makes it small and light enough to be used for animal behaviour purposes, even on small animals.
    The device runs with an ESP32 Pico D4 microcontroller unit whose main CPU has a 240MHz maximum clock speed, $4\si{\mega\byte}$ flash memory and $520\si{\kilo\byte}$ RAM.
    The WildFi can store up to $256\si{\mega\byte}$ of sensor readings in its NAND flash memory.
    It is equipped with a Bosch BMX160~\cite{bmx160} 9-axis inertial measurement unit (IMU), thus including a 3-axis accelerometer, gyroscope, and magnetometer, allowing for the exploration of the impact of sensor fusion for our purposes.
    The recording frequency of the IMU is 50Hz, and the data it records lies at $900\si{\byte}/s$.
    It also has a natively integrated environmental sensor, the Bosch BME680~\cite{bme680}, with a sampling frequency of 1Hz and a recording rate of $10\si{\byte}/s$.
    The WiFi transmission rate of the WildFi lies at $230\si{\kilo\byte}/s$ with an average current consumption of $108mA$, as measured by the authors~\cite{Wild2022}.
    
    Since the purpose of this work is to save energy on the bio-logger, we only executed the data acquisition and the final classification model on the bio-logger.
    We outsourced the modelling process to a tower PC.
    The PC used for the training had an 11th Gen Intel(R) Core(TM) i7-11850H 64-bit processor, 2.5GHz dual-core, and 32GB RAM.
    We used Python 3 and the SKlearn library~\cite{pedregosa2011scikit} for generating our machine-learning models.

    \subsection{General Concepts}   
\begin{figure}[h!]
        \centering
        \includegraphics[width=\textwidth]{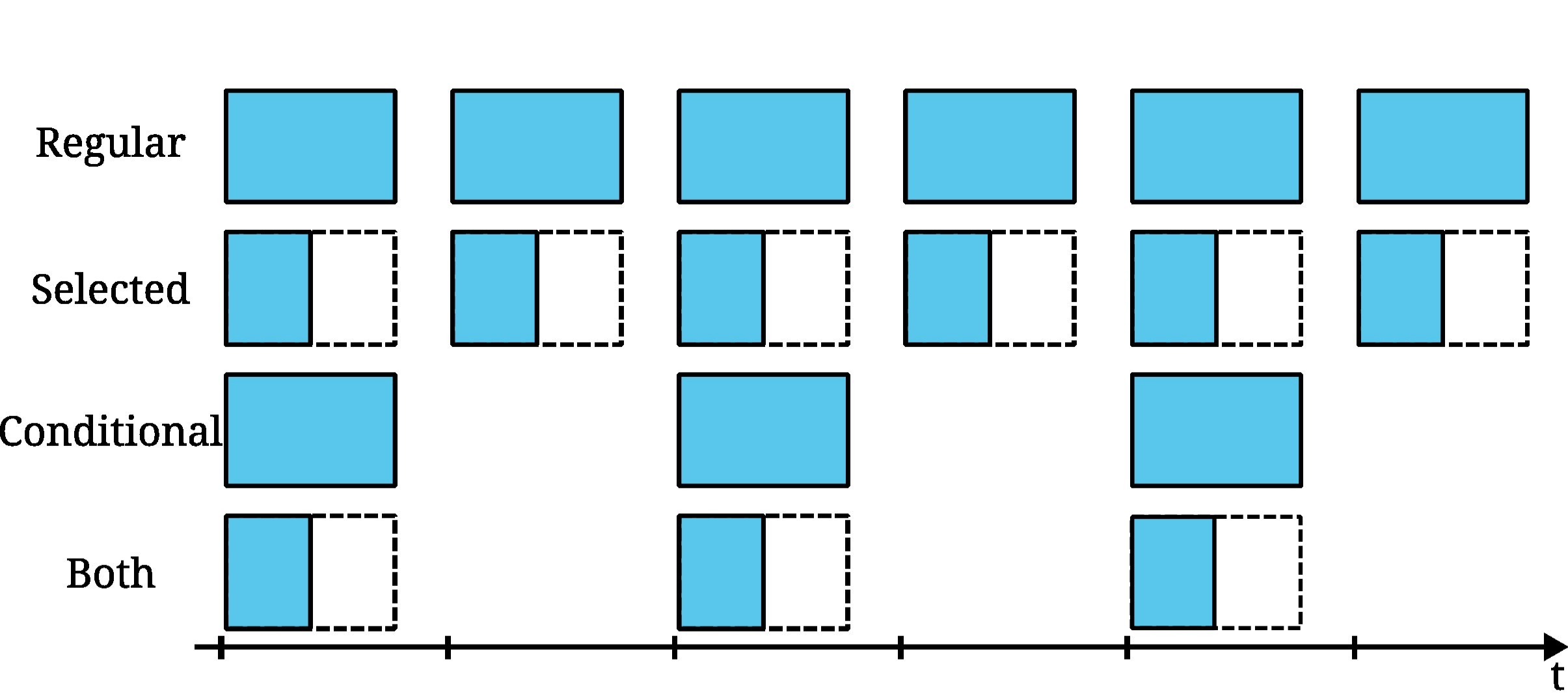}
        \caption{This image illustrates the concepts behind the reduction of transmission time with the help of contextual transmission. "Regular" means to send every message whole. "Selected" means to send only a selection of information. "Conditional" means to send messages when a condition is met. "Both" corresponds to a combination of conditional and selected data transmission. Each blue box indicates a message that is being sent. The dotted lines illustrate that a portion of a message is not transmitted. The bottom arc shows a progression over time, without a specified unit of time, to allow for an abstract comparability between the different approaches.}
        \label{fig:allofit}
    \end{figure}

    Commonly, energy consumption can be described as the product of the required power in watts multiplied by the time the system is running.
    It is constituted by $E=P\cdot t$ or directly through the respective units $J=W\cdot s$.
    Therefore, to reduce energy consumption, we can reduce either the wattage required for an action or the time $t$ invested in that action.
    We focus on reducing $t$ by reducing the total time required for data transmission, as we can control this aspect more easily with software solutions.
    The time required for a transmission $T$ can be calculated with $T=\frac{L}{R}$, with $L$ being the length of a message and $R$ being the transmission rate of the transceiver.
    The calculation is simplified due to different overheads, like connection time, but suffices to explain the concepts in our context.
    For the remainder of this work, for simplicity, we will consider the following assumptions based on the original WildFi publication:
    Firstly, we assume an invariant data transmission rate $R=230\si{\kilo\byte}/s$.    
    Secondly, we neglect overheads for transmission time, as they are included in the energy costs the original authors show.
    Also, we will assume a supply voltage of $3.75V$.
    Lastly, we additionally assume that $T=t$, meaning that the time over which energy is consumed directly corresponds to the transmission time.
    To calculate the energy consumption we derive $P=3.75V\cdot 0.108A=0.405W$, which leaves us with the following equation to calculate energy expenditure for data transmission for the WildFi tag: $$E=\frac{0.405W\cdot L}{2.3\cdot 10^5\si{\byte}/s}$$
    As a result, in our setup, the reduction of energy expenditure directly corresponds to reducing $L$.
    To reduce the time there are multiple options as Fig.~\ref{fig:allofit} illustrates.

    Given a set of messages with a uniform length $L$ per message, reducing the overall time of transmission can be achieved by either reducing the number of messages sent, reducing $L$ of each message, or both.
    A reduction of the number of transmitted messages can be achieved through conditional data transmission, where a bio-logger sends data only when specific patterns are detected.
    This approach minimizes network traffic while preserving the integrity of each message and relies on recognizing patterns we will refer to as behaviours of animals.
    For a bio-logger that can detect a behaviour $B$ with $100\%$ accuracy, data transmission can be limited to occurrences of $B$, reducing messages proportionally its frequency.
    Alternatively, data can be transmitted when $B$ is not detected, though this risks omitting relevant information if behaviours overlap.
    Ideally, behaviours of interest and $B$ should be mutually exclusive in this scenario.
    In practice, perfect detection is rare.
    Therefore, the goal is to minimize the loss of relevant data.
    For instance, if a classifier misclassifies $1\%$ of the time, and the target behaviour is common, such as when an animal is sleeping, missing a few instances may be acceptable.

    If maintaining the density of data points is essential, reducing the message length $L$ is an alternative.
    This reduction can be achieved through selected data transmission, where decisions are made about which values, such as sensor data or their encodings, to include in each message.
    This process involves either encoding or compression techniques, as discussed in prior research~\cite{jain2018gist,lelewer1987data}.
    While encoding preserves information, data compression may introduce some distortion.
    Both methods aim to reduce message size while retaining the core information.

    Generally, each data transmission protocol benefits from the reduction of data to transmit, if preserving energy is the goal. 
    Different data transmission protocols have their respective limitations and advantages, however.
    While the SigFox protocol allows only about 140 messages with 12$\si{\byte}$ each, its transmission range can reach distances of several kilometres~\cite{wild2023multi}.
    On the other hand, the  WiFi protocol of the WildFi tag can transmit vastly more data than the SigFox, with a transmission range of only a few hundred meters.
    In this study, we consider the WiFi protocol of the WildFi tag and an example scenario where the transmitted data are the features required for behaviour detection.
    These features might include direct sensor values or proxy methods from literature, such as VeDBA~\cite{gleiss2011,Lopez2021}, which encodes all three acceleration axes into a single value correlating with energy expenditure.
    Such approaches preserve key information while reducing the size of the data transmitted from the bio-logger to the receiver.

    \subsection{Pattern Recognition Procedure}
\begin{figure*}[h!]
        \centering
        \includegraphics[width=\textwidth]{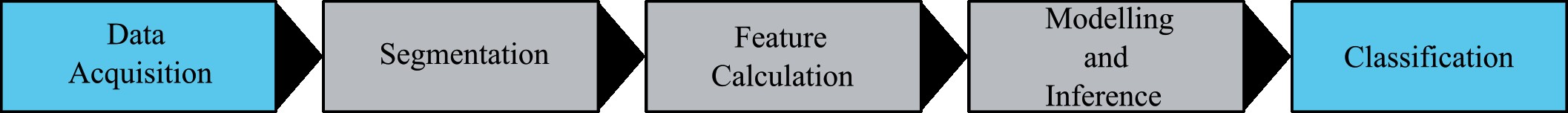}
        \caption{This is a modified version of the activity recognition chain pipeline by Bulling et al.~\cite{bulling2014tutorial}. From left to right, the single boxes show the abstract steps that are appropriate to achieve a successful classification. We additionally marked which step happens on which type of device, either PC or bio-logger, by colouring the PC steps grey and the onboard steps blue. The original author permitted us to use their graphic.}
        \label{fig:arcsymb}
    \end{figure*}

    To produce a model we utilised the framework for effective pattern recognition from recording sensor data to executing a classification called the "activity recognition chain" (ARC) by Bulling et al.~\cite{bulling2014tutorial}.
    We applied it as the frame of our workflow to establish a well-accepted starting point.
    As the simplified version of that ARC in Fig.~\ref{fig:arcsymb} shows, there are 5 main processing steps.

    \subsubsection{Data Acquisition}
        The data acquisition corresponds to the actual measuring by sensors.
        These sensors can be anything from acceleration sensors to GPS or humidity sensors.
        We acquired our data by recording it using the WildFi tag.
        The activated sensors for measuring were the IMU with accelerometer, magnetometer, and gyroscope, as well as the environmental, and the GPS sensor.
        The latter was attached and active for technical reasons, but was not considered in further processing.
        In general, the logger was configured to read data 50Hz IMU and 1HZ GPS data.
        The recorded raw data was then decoded and converted into CSV file format for further processing.
        The resulting file represents a time series where each row is a timestep.
        All timesteps are 1 second apart, and each row is a n-tuple with every measurement from its respective moment, including the timestamp and 50 values per axis of the accelerometer, the gyroscope, and the magnetometer.
        When we refer to a data point, we refer to one of the n-tuples.

    \subsubsection{Segmentation}
        The segmentation step is the data preprocessing step. 
        For this step, we removed any entailing readings that did not belong to our test case and labelled the remaining data afterwards.
        Since the IMU samples with 50Hz bursts per sensor, we took the average of all 50 readings per second as the corresponding value for the respective sensor.
        For tracking animal behaviour this is likely to smooth out information which is crucial to detect complex movement patterns, but for our experimental data, this approach sufficed.

    \subsubsection{Feature Calculation}
        Feature calculation is a preparatory step to provide features which are capable of distinctively describing the target behaviour.
        The question of which features in the form of sensor values are especially descriptive of a given behaviour of an animal proves particularly difficult.
        Too many features for training will result in high execution times for the modelling.
        Too few might not capture an animal's behaviour sufficiently.

        Table~\ref{tab:williams} shows a small derivative version of Tab. 1 from the work of Williams et al.~\cite{Williams2019}, who put into perspective different types of sensors and in which context they suit best.
        We incorporated this part of their work as part of our feature calculation step since it helps to rule out potentially irrelevant sensors given knowledge of the context of the usage of the bio-logger.
        Various analysis techniques can and should be utilised to filter features further if the actual feature space amounts to an unfeasibly long time for modelling.
        
        As we pursued the goal of putting the abilities of sensors into perspective, we limited our relevant sensor types to intrinsic ones, namely the accelerometer and the gyroscope, omitting the magnetometer for simplicity.
        Thus, also our feature space is limited to the three axes of the accelerometer (AX, AY, AZ) and the gyroscope (GX, GY, GZ), amounting to a total of 6 input features.
        
        \begin{table}[h!]
            \centering
            \caption{Small overview of sensor types and their field of use}
            \begin{tabular}{|l|l|l|l|}
            \hline
            \textbf{Sensor type} & \textbf{Examples}        & \textbf{Description}        & \textbf{Relevant}           \\ 
                                 &                          &                             & \textbf{Questions}          \\ \hline
            Location             & GPS                      & Location-based              & Space use;                  \\
                                 &                          &                             & interactions                \\ \hline
            Intrinsic            & Accelerometer;           & Patterns in                 & Behavioural                 \\ 
                                 & Gyroscope; Magnetometer  & body posture                & identification              \\ \hline
            Environment          & Temperature              & Record external             & External factors            \\ 
                                 &                          & environment                 &                             \\ \hline
            \end{tabular}

        \label{tab:williams}
        \footnotetext{A reduced version of Table 1 from the work of Williams et al.~\cite{Williams2019}, where the authors break down in which cases which sensors are appropriate. The "Sensor Type" column shows the possible contexts of sensors. The column "Examples" covers which sensors are fit for the sensor types. The "Description" column indicates the kind of information that can be expected and "Relevant Questions" hints at what research questions can be addressed with the respective sensors.}
        \end{table}

    \subsubsection{Modelling and Inference}
        The Modelling and Inference step describes the procedure of the actual training of the model.
        There are methods for so-called tinyML, machine learning on microcontrollers~\cite{dutta2021tinyml}, which can require as little as $16\si{\kilo\byte}$ RAM, like TensorFlow Lite by Tensorflow~\cite{tensorflow2015-whitepaper}, which easily fits into the WildFis memory.
        To conserve energy, we outsource this process to a PC.
        We chose decision trees~\cite{kingsford2008decision} as our model due to their hierarchical structure, which helps identify the most impactful features for describing behaviour.
        Unlike other machine learning approaches, decision trees require only a few hyperparameters to be tuned before the actual training.
        The primary parameter in our case is $k$, which sets the maximum tree depth, allowing us to limit the computational steps during the classification process on the bio-logger and, thus, actively influence the current consumption during the process.
        However, we outsource this process to a PC to preserve energy.
       
        In machine learning the term hyperparameter tuning refers to the act of adapting the hyperparameters and retraining the model with the new hyperparameter to evaluate the quality of the different models to pick the hyperparameters which return the model that fulfil the desired qualities the best. 
        After a few iterations with our data, we obtained $k=7$, as a well-performing tree depth.
        We also worked with $k=14$ according to a related project preceding this study, where this was found to be the maximum depth for optimal usage of the WildFi tag.
        As for the models, we pursued two different directions.
        The first direction was to use all features for a decision tree, whose purpose is to investigate the overall functionality of decision trees for behaviour detection, as well as the functionality of the deployed code.
        We trained the model with all three axes of the accelerometer and the gyroscope and deployed the resulting decision tree on the WildFi tag.
        
        The second direction introduces the selected data transmission.
        We applied the hyperparameter tuning approach to the feature space such that we obtained a model for each permutation of the desired training features.
        We checked how well the decision trees perform in terms of F1 score, the harmonic mean between precision and recall of a classifier, and accuracy for each of the resulting models.
        Afterwards, we compared the different scores and were able to pick a feature setup that can be considered good enough for the recognition of a chosen behaviour.
        In the end, we had two decision trees, one with all and one with a subset of features.

    \subsubsection{Classification}
        The classification of real-time readings of data happens on the bio-logger as this is where we want to achieve energy savings.
        For this, a working classifier has to be deployed on said bio-logger.
        We implemented a small Python program, which produces the decision trees with the help of the recorded data and outputs them as header files which can be included by the WildFi. 
        With a small modification of the logger's firmware, it is then able to use the produced tree.
        However, the on-board classification introduces additional calculations required for the preparation of classification or the classification itself, which are not allowed to be more expensive than the savings achieved.
        To estimate the costs of an operation, we assume that the WildFi runs at its most power-consuming mode with 240MHz clock speed.
        According to its datasheet, the ESP32, at 240 MHz, requires about one clock cycle for a simple operation like addition or comparison, which lies at $\frac{1}{240\times 10^6}Hz=4.17ns$.
        Say we apply an operation with 100 such calculations on a sensor reading such that an entire operation takes $417ns$.
        Assuming 10 sensors, this ends in on average $4170ns$ per sampling. 
        The actual consumption can be roughly calculated with $P=C\times V^2 \times f$~\cite{rabaey2002digital}, where C is the capacitance, $V=3.75$, and $f$ is the clock frequency.
        Since C varies, we use active power consumption as a reference. 
        The active current is, therefore, $P=0.24A\cdot 3.75V=0.9W$.
        Thus, with $E=P\cdot t=0.9W\cdot 4170\cdot 10^{-9}s=3753\cdot 10^{-9}J=3.753\cdot 10^{-6}J$.
        Now say we transmit the smallest information we can with only 2 bytes per sensor per second with the assumed 10 sensors, thus $20\si{\byte}/s$.
        We can calculate the energy costs with $E=\frac{0.405W\cdot L}{2.3\cdot10^5\si{\byte}/s}=\frac{0.405W\cdot 20}{2.3\cdot10^5\si{\byte}/s} =3.52\cdot 10^{-5}J$, which is an order of magnitude more than the calculations done beforehand, even in a scenario, where the transmitted information would be small.
        If we use a more expensive operation, like division with approximately 32 clock cycles we arrive at $\approx 1.2\cdot 10^{-5}J$, which still is almost a third of the cost of transmission. 
        Therefore, as long as the on-board calculations are not extremely extensive like a larger number of trigonometric calculations or indefinite calculation cycles, the on-board classification is cheaper than just transmitting everything.

    \subsection{Human-based Experiment}
    Since collecting and annotating animal-borne data is challenging, we conducted an experiment to detect the body movements of humans.
    We wanted to be able to read basic movement types from a labelled data set, train a classifier and deploy it on the WildFi tag, thus teaching it to recognise one specified type of movement.
    We chose 5 behaviours: lying, sitting, standing, walking, and running.
    
    We had 3 people whose movements we tracked with the WildFi tags.
    We will further refer to the three tags as EA60, EBF8, and ED3C which are the last four digits of their respective IDs.
    Each of the devices was encased inside a tight-fit casing, which in turn was attached centrally to a baseball cap's brim.
    To record the data, all 3 participants wore these baseball caps.
    Thus, to refer to participants 1,2 or 3 we can also refer to the respective tag IDs (Tab.~\ref{tab:distribution}).
    The participants had to fulfil tasks which encompassed the 5 different movement types.
    Each task had a duration of an approximate multiple of a minute and the whole recording had a duration of about 31 minutes.
    Between each two tasks, a short transition phase took place, which has varying lengths due to individual differences in movement habits like how long it takes an individual to stand up, for example.
    The procedure was video recorded to reduce outliers for the training data by assisting the labelling process.

    The produced data sets were then decoded and post-processed to remove measurements which did not lie inside the period of the experiment, as well as the transition times between two target behaviours.
    Finally, they were labelled using the video recording.
    
    For the generation of the decision trees, we chose a total of 8 features to train with.
    All six training features stemmed from the IMU, specifically the accelerometer and the gyroscope.
    From the accelerometer, we used the mean per reading of all three axes (AX, AY, AZ), as well as the calculated VeDBA score as a well-established metric.
    From the gyroscope, we used the variance per reading of all three axes (GX, GY, GZ), as well as the VeDBA calculation function applied on the axes, which we named GVeDBA.
    The latter is no common metric and we were experimenting with it as a sort of energy expenditure equivalent from rotation instead of translation. 
    We kept it included in our feature space since it produced promising results consistently.
    We generated decision trees for each of the participants, with tree depths 7 and 14, based on their respective data and compared the trees with all available features to those with only a subset of features by evaluating their confusion matrices.
    The behaviour we targeted to optimise was "standing".
    Apart from the decision trees and confusion matrices for all trees, we also produced rankings for the different permutations of features based on their performance for recognising the targeted behaviour.
    Lastly, we estimated potential savings in transmission time based on the outcomes.

\section{Results from human-based experiment}\label{sec2}
    
\begin{figure*}
        \centering
        \includegraphics[width=\linewidth]{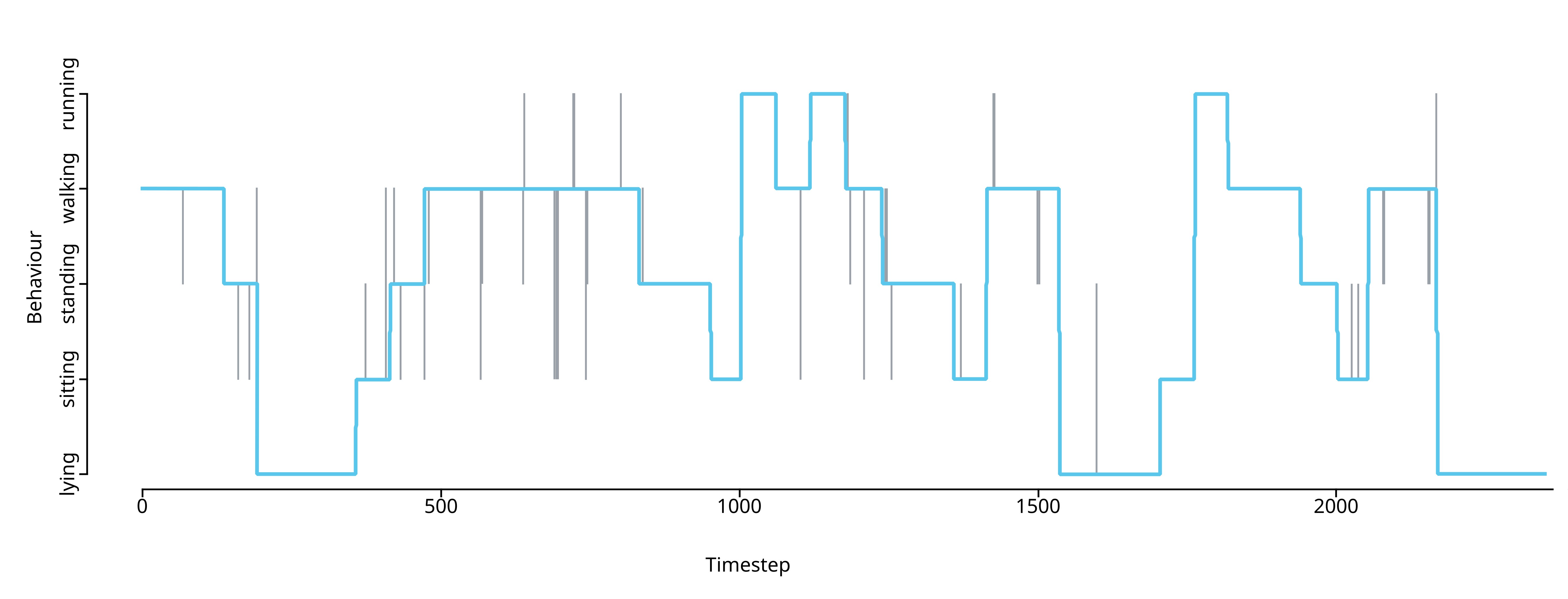}
        \caption{This plot compares the actual behaviours with the classified behaviours throughout the experimental data. The y-axis indicates the different behaviours. The x-axis shows the timesteps. The blue line shows which behaviour was exhibited during the recording, therefore mapping to the actual labels of the time series data. The grey lines connect the behaviour classified by the decision tree with the actual behaviour at that particular timestep. Therefore, the grey lines indicate wrong classifications.}
        \label{fig:verlauf}
    \end{figure*}

    The datasets we produced, after preprocessing, have 2350 data points for EA60, 2320 for EBF8, and 2240 for ED3C. 
    The overall distribution of the 5 behaviours per individual can be seen in Tab.~\ref{tab:distribution}, with walking being the majority and running the minority class.
    The approximate data size per individual stems from the number of data points multiplied by the sampling rate of the tri-axis accelerometer and gyroscope, thus, $6\cdot 50Hz\cdot 2\si{\byte} \cdot \#timesteps$.
    Figure~\ref{fig:verlauf} illustrates the different phases of the underlying time series, by mapping the respective behaviours (y-axis) to the points in time when they happened (x-axis) with a blue line.

    \begin{table}[h!]
        \caption{\label{tab:distribution} Data overview per sensor}
        \begin{tabular}{|l||lll|}
            \hline
            Tag ID & \multicolumn{1}{l|}{\begin{tabular}[c]{@{}l@{}} EA60\end{tabular}} & \multicolumn{1}{l|}{\begin{tabular}[c]{@{}l@{}} EBF8\end{tabular}} & \multicolumn{1}{l|}{\begin{tabular}[c]{@{}l@{}} ED3C\end{tabular}} \\ \hline \hline
            Participant    & $1$ & $2$ & $3$ \\ \hline \hline
            \# Data points    & $2350$ & $2320$ & $2340$ \\ \hline \hline
            $\approx$ Data size    & $1410\si{\kilo\byte}$ & $1392\si{\kilo\byte}$ & $1404\si{\kilo\byte}$ \\ \hline \hline
            \% Lying      & $21.97$ & $21.3$  & $21.98$ \\ \hline
            \% Sitting   & $11.58$ & $11.82$ & $11.93$ \\ \hline
            \% Standing    & $17.62$ & $17.9$  & $12.44$ \\ \hline
            \% Walking    & $41.51$ & $41.57$ & $46.47$ \\ \hline 
            \% Running    & $7.32 $ & $7.42$  & $7.18$  \\ \hline
        \end{tabular}
        \footnotetext{This table shows the distribution of the behaviours in each training data set for the three individuals proportionately. It displays which participant had which sensor and how many data points the respective time series' had after processing. The number of data points is equivalent to the number of time steps.}
    \end{table}

    The training of the decision trees resulted in 24 different trees for comparison (see tab.~\ref{tab:comparison}) for which we also produced header files which can be used by the WildFi to classify the behaviours on-board directly by using the sensor values.
    The full feature set consistently over individuals and tree depths shows a higher F1 score than accuracy, while the feature subsets consistently have higher accuracy than the F1 score.
    Also, each tree depth 14 version surpasses the tree depth 7 version in terms of F1 and accuracy.
    Furthermore, for each individual, it holds that a subset of features holds a higher accuracy compared to the full feature set while showing a lower F1 score.
    Notably, the feature subset for Participant 3 and a tree depth of 7 completely outperformed the full feature set.   
    
    \begin{table}
            \centering
        \caption{Decision tree performances by sensor and tree depth}
        \label{tab:comparison}
        \begin{tabular}{|l|l|l|l|l|}
        \hline
        Tag ID \&  & Full Set& Full Set & Subset & Subset  \\
        Tree depth & F1& accuracy& F1& accuracy\\ \hline
        EA60 TD  7  &    79.37  \%  & 65.90 \% & 66.01 \%   & 79.46 \%\\ \hline
        EA60 TD 14  &    92.34  \%  & 85.81 \% & 87.24 \%   & 93.16 \%\\ \hline
        EBF8 TD  7  &    73.33  \%  & 58.01 \% & 60.02 \%   & 74.93 \%\\ \hline
        EBF8 TD 14  &    90.04  \%  & 81.94 \% & 83.22 \%   & 90.80 \%\\ \hline
        ED3C TD  7  &    62.62  \%  & 45.74 \% & 69.56  \%  & 82.00 \%\\ \hline
        ED3C TD 14  &    96.34  \%  & 92.96 \% & 93.22  \%  & 96.48 \%\\ \hline
        \end{tabular}
        \footnotetext{A table comparing the F1 scores and accuracies between the full feature set and the subset of features for the decision trees per device. For the subsets, only the best-performing feature permutations are covered here. For all trees used for comparison, the tree depths of 7 and 14 were chosen. The behaviour for which they were optimised was "standing". The first column has the tag IDs together with their tree depth (TD).}
    \end{table}
    
    Table~\ref{tab:interesting-permutations} shows some permutations of EA60 with a tree depth of 14 together with their ranking in terms of quality of classification for the behaviour "standing".
    The first finding to notice is, that we obtain a higher accuracy as well as a higher F1 ratio while not using all available features.
    In fact, the decision tree which uses all features ranked in the twentieth place.
    Also noteworthy is, that the gyroscope is included with at least one axis in the first 200 of 256 configurations.
    Rank 38 is the first permutation which only requires 3 features and still offers a satisfying accuracy of $90.47\%$. 
    A last observation is that rank 3 is less than $1\%$ less accurate than rank 1, but requires a whole axis less to achieve this, which is $20\%$ less relevant data to record and or transmit. 

    \begin{table}[h!]
    \centering
        \caption{Feature permutations ranked by the chosen quality metrics}
        \label{tab:interesting-permutations}
        \begin{tabular}{|l|l|l|l|}
            \hline
            n-th      & Feature Permutation             & F1      & Accuracy     \\ 
            best      &                                 & in \%   & in \%         \\ \hline
            1         & GX;GY;GZ;AX;AZ;                 & 87.24   & 93.16         \\ \hline
            2         & GX;GY;GZ;AX;AZ;                 & 87.04   & 93.04         \\ 
                      & GVeDBA;                         &         &               \\ \hline
            3         & GX;GY;GZ;AZ;                    & 85.95   & 92.41         \\ \hline
            4         & GY;GZ;AX;AY;                    & 85.31   & 92.04         \\ \hline
            5         & GY;GZ;AY;VeDBA;                 & 85.02   & 91.87         \\ \hline
            ...       & ...                             & ...     & ...           \\ \hline
            20        & GX;GY;GZ;AX;AY;                 & 83.81   & 91.16         \\ 
                      & AZ;VeDBA;GVeDBA;                &         &               \\ \hline
            ...       & ...                             & ...     & ...           \\ \hline
            38        & GX;GZ;AX;                       & 82.65   & 90.47         \\ \hline
            ...       & ...                             & ...     & ...           \\ \hline
        \end{tabular}
        \footnotetext{This table shows a selection of feature permutations for EA60 with a tree depth of 14. The "n-th best" column signifies the rank of the respective feature permutation regarding its quality measures for classifying a priorly specified behaviour. In this case, this behaviour is "standing".}
    \end{table}

\begin{figure}
        \centering
        \includegraphics[width=\linewidth]{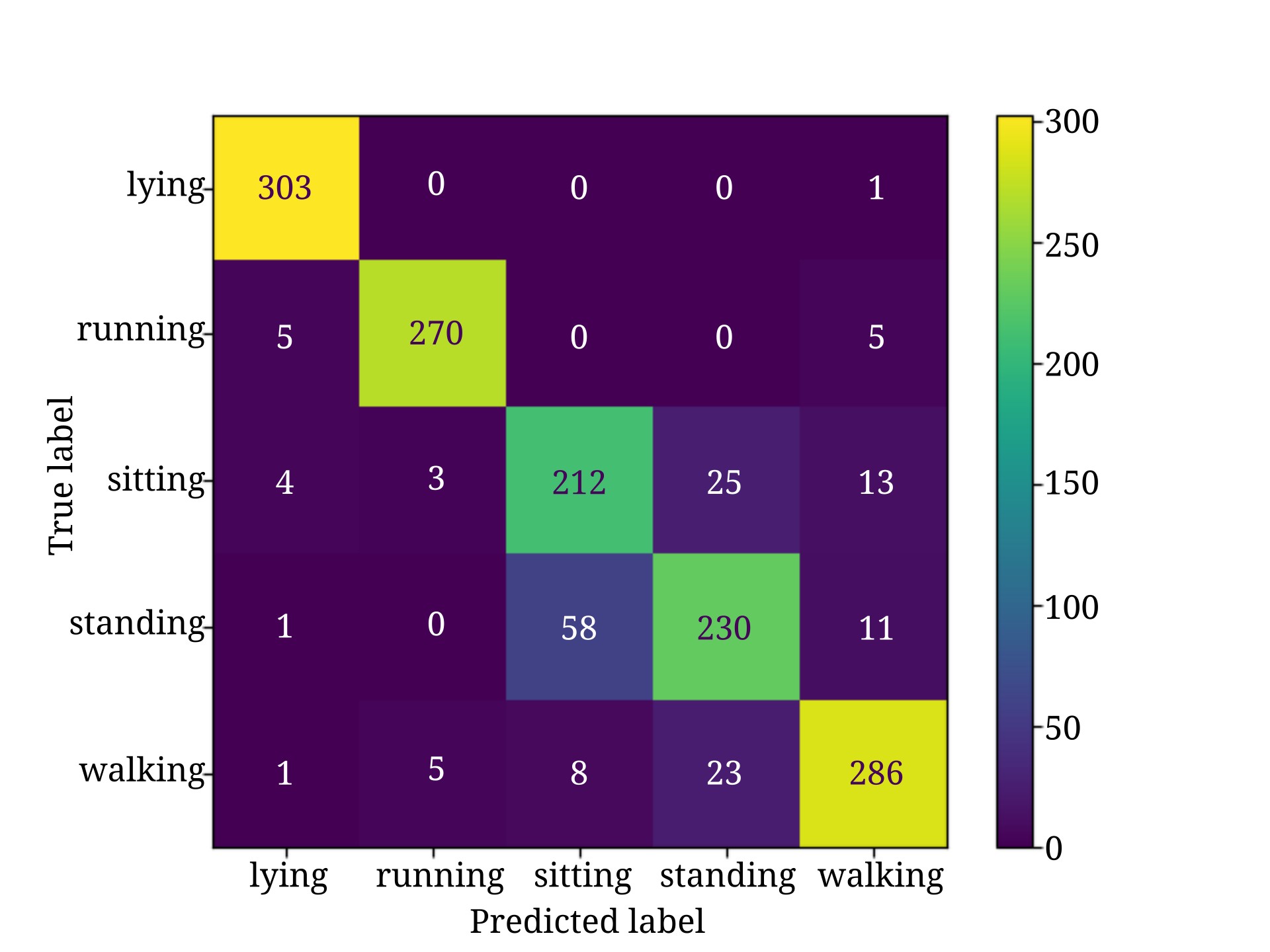}
        \caption{This image shows the confusion matrix for the decision tree, optimised for the behaviour standing, with a tree depth of 7, belonging to Participant 1. The decision tree has an increased number of misclassifications between sitting, standing and walking. The y-axis shows the actual behaviours and the x-axis what the classifier has predicted.}
        \label{fig:confusion}
    \end{figure}

    To illustrate the actual performance of the decision tree, we exemplarily illustrate the classification of the EA60 tree using a feature subset and a tree depth of 14 in Fig.~\ref{fig:verlauf}.
    There, the grey lines connect the classified behaviour and the actual behaviour at that time step, such that it becomes visible when and how often the classifier got its classification wrong.
    It quickly becomes visible that the classifier had the most problems with seeing the difference between sitting, standing and walking.
    The respective confusion matrix in Fig.~\ref{fig:confusion} shows that this problem becomes even more evident for a tree depth of 7, especially for the differentiation between standing and walking.
    Applying the classifier with a tree depth of 14 for EA60 on ED3C results in an accuracy of $27.13\%$ and $25.44\%$ for a tree depth of 7.

    Since there are only a few false negatives for "standing", we now assume the classifier is feasible to detect that behaviour and we can conduct some example calculations for energy saving for the WildFi tag.
    For conditional data transmission, taking Table~\ref{tab:distribution} as a basis and considering "standing" to occur with an estimated frequency of $17.62\%$ for EA60, the WildFi tag would now be able to transmit information with an estimated frequency of $17.62\%$.
    Therefore, only $2350\cdot 0.1762\approx 414$ data points of the time series would be appointed for saving and transmission, reducing the overall message length from $1410\si{\kilo\byte}$ to $248.4\si{\kilo\byte}$ and the consumed energy from $662mA$ to $116.64mA$. 
    
    As selected data transmission is versatile, for our calculation, we will assume that we only transmit the information which leads to the classification.
    In that case and concerning Table~\ref{tab:interesting-permutations}, the solution with the highest accuracy would exclude one axis of the accelerometer readings.
    Since the initial amount of input features suggested six IMU features, which record with a rate of $600\si{\byte}/s$, the reduction of information to transmit of the IMU lies at $\frac{1}{6}$.
    The approximate data size for EA60 in the given scenario reduces to $1175\si{\kilo\byte}$, accordingly, with a reduction of power consumption to $\approx 551.74mA$.
    
    With conditional and selective we reach a total reduction to $0.1762\cdot \frac{500}{600}\approx 0.1468=14.68\%$.
    Thus, we only transmit $14.68\%$ of the amount of data compared to transmitting all of the raw data from the IMU, which equals a consumption of $97.2mA$.
    The second, more extreme, example is to only send a $2\si{\byte}$ sized signal (using the unsigned short data type), the timestamps for when a target behaviour is detected.
    In that case, the total transmitted data is reduced to 2 times the frequency of a behaviour times the number of data points.
    For our example that would be $2\cdot 0.1762\cdot 2350\approx 828\si{\byte}$ such that the final transmission costs are reduced by $99.9903\%$.

\section{Discussion}
    The main goal of this study is to investigate how to utilise machine learning for reduced energy consumption in the context of data transmission on animal-borne devices.
    To approach this problem, we simplified the energy consumption of data transmission to directly relate to the transmission time, omitting energy consumption overhead caused by different aspects, such as the time required for the transceivers to connect.
    To shorten transmission time we in turn aimed for a reduction of the overall amount of data transmitted.
    We argued, that there are several ways to achieve this, like data compression, encoding, or filtering.
    In this study, we concentrated on the filtering aspect and partly utilised encoding, but without an emphasis on it.
    Consequently, in future work, more emphasis on data compression and encoding is required.
    Filtering, as we argued, can happen in the form of either filtering out whole data points, parts of each data point, or a combination of both.
    We named the two forms of filtering conditional and selective data transmission, since for the former we have to decide under which circumstance to transmit and for the latter what data exactly to transmit.
    To make decisions about what to filter, we used machine learning in the form of pattern recognition.
    Specifically, certain states of an animal, namely their behaviours, could aid in making a distinction between what to send, or store, and what not.
    Since the identification and classification of animal behaviours from time series data has gained popularity in the younger past, also due to the increased availability of well-performing bio-loggers, we focused on that.

    To achieve classifications, as we wanted to study the potential behind machine learning for energy preservation, we decided to use decision trees.
    Their structure, which is very simple compared to more elaborate models like deep neural networks, allows us to control the number of computations per classification directly and to interpret what exactly impacted their quality.
    Another benefit is that decision trees, due to the way they work, can be translated into simple pieces of code.
    We made use of this fact and derived code files for the classification models we obtained that we were able to deploy on the bio-logger which we used - the WildFi tag~\cite{Wild2022}.
    Thanks to the high computational capabilities of these state-of-the-art bio-loggers we can therefore do classification on the bio-logger, which in itself presents a huge advancement to the field of bio-logging.
    However, decision trees are not nearly as powerful as other machine learning models concerning recognising behaviours so further investigation on how more complex classifiers could be used on devices like the WildFi tag is required.
    
    To produce pattern recognition models, from the computer science perspective, we were guided by the activity recognition chain proposed by Bulling et al.~\cite{bulling2014tutorial}, such that our general approach has a sound foundation.
    Due to our goal to preserve energy, we outsourced as many steps as possible from the loggers to a PC.
    Thus, although possible, we did not train our decision trees on the WildFi.
    Furthermore, we incorporated parts of the work of Williams et al.~\cite{Williams2019} about interdisciplinary work on animal behaviour to account for the complexity of animal behaviour-related topics.
    Through the integration of their findings about which sensor types are relevant for which kind of question the experiment we made to gain further insights gained more validity. 
    As an easy proof of concept, we conducted a small experiment by recording human data, applying our workflow to it, analysing the results and inferring the practicality of it, as well as gaining some numbers to do example calculations.
    The data we recorded consists of roughly 30 minutes of data points per participant and mirrors five different and distinct behaviours.
    We chose the behaviours, such that they would all be patterns in body posture, hence requiring intrinsic sensor types. 
    Furthermore, they were all mutually exclusive for the most part, thus simplifying the data set.
    As a consequence, our experiment has limitations in the sense that we do not fully depict a realistic scenario where different behaviours might overlap.
    A cat might simultaneously walk and chew, for instance.
    However, since we only want to investigate the potential of machine learning in this study, we deliberately designed this simpler use case.

    To gain an even deeper insight, we did not only look into general decision trees, but we trained a number of them with different depths and all possible feature combinations to see, which sensor may have more or less impact on successfully identifying behaviours.
    Our experimental results suggest, that the decision trees with a maximum tree depth of 14 are capable of classifying with a high accuracy.
    This was to be expected since 14 is likely an overfitting.
    However, tree depths of 7 still maintain relatively high accuracy with $82\%$ for ED3C, for example.
    These outcomes can likely be further improved with different features or more data than only half an hour of data points or by employing more sophisticated sampling methods than oversampling.
    Also, data from more sources could help to overcome the inability to use a classifier inter-individually.
    If we had data from a larger number of individuals with the same quality, the individuals' respective movement idiosyncrasies, like habit-, gender- or age-dependant nuances, would not impact the classifier as much.
    We might have been able to use the resulting classifier on a person who did not participate in the recording with higher accuracy than the deficient accuracy we had between EA60 and ED3C.
    Generally, not only do different individuals have their movement habits, but most of the different species likely have vastly different patterns for the same behaviour, necessitating species-specific experimentation, utilising the general approach we present.  
    
    Another interesting finding has to do with the feature space.
    Specifically, the gyroscope proves to be important for behavioural recognition.
    It appears in all of the best-performing configurations of our feature space for "standing".
    An explanation could be, that while an individual is idly standing, it will move its head around more than during a running action.
    And since the bio-logger was effectively attached to the head, this would result in more changes in rotation and less in acceleration, which the classifier would use to recognise the behaviour.
    Generally, some systematic and context-dependant head movement is likely linked to certain behaviours, which the gyroscope captures in a different granularity than the accelerometer.
    This appears to help in distinction, thus rendering the gyroscope important for classification purposes.
    Apart from the exact quality metrics, it becomes clear, that classifiers that work for individuals on bio-loggers appear to be possible with only about 30 minutes' worth of data.
    How long a time series has to be in the end depends on the nature of the targeted behaviour, external influences and further factors, however.
    
    Under the assumption of a classifier which could predict "standing" reliably, we did some example calculations for possible energy savings based on a hypothetical scenario, where the stored and transmitted data is limited to the data points required to detect a given target behaviour.
    We predicted, that the transmission time for sending only occurrences of the behaviour "standing" together with 5 features would be reduced to approximately $14.68\%$.
    The actual impact on the overall runtime of the WildFi tag depends on the setup, however.
    Assuming Tab.~2 from the original publication~\cite{Wild2022}, the increase in energy consumption ranges from as little as $0.03\%$ to $58\%$.
    A similar reduction of the $58\%$ would leave us with only $8.51\%$ extra cost through transmission, thereby increasing the runtime of the device from 94 to about 137 days, giving researchers over a month of extra time to collect relevant data.
    What is more is, that the higher the impact of the transmission, the higher the gain of smart choices in what to transmit is.
    A big limitation here is, that there are a lot of circumstances which influence the results.
    For one, classifiers do not offer 100 per cent solutions.
    Their reliability depends on the data, the actual behaviour of the animal they operate on, the way the classifier model was created and more.
    Also, the software and hardware setup of a bio-logger determines, which impact such machine-learning solutions can have and how extensive they can be.
    For example, we discussed the approximate energy expenditure of a single operation.
    These costs highly influence how extensive operations on a bio-logger can be before they exceed the costs for transmitting the raw information and directly depend on the used hardware and its configuration.
    
    Despite the various factors of influence and even though omitting certain sets of data points can also be achieved through other means than machine learning, the possibility of using classifiers as an onboard solution offers more options since elaborate contextual decisions can be achieved.
    For example, bio-loggers could react in situ and send warnings if an animal exhibits unusual behaviour, enabling responsible persons or scientists to react timely.
    Also, such classification approaches could be used to govern the very limited transmission capabilities of devices that use SigFox as a transmission protocol.
    By that, the high transmission range could be utilised much more efficiently. 
    With these insights, we combined the works of several researchers, such as Korpela et al., Bulling et al., and Williams et al.
    We showed, that given a capable bio-logger with WiFi technology, machine learning is not only possible on such a device, but is also feasible for significant current consumption reduction through smart decision-making on when to transmit which information.

\section{Conclusions}\label{sec13}
    In this study, we explored the potential of using machine learning, specifically decision trees, to optimize energy consumption in bio-logging devices through intelligent data transmission.
    We did this exemplarily through the use of on-board classification on the WildFi bio-logger.
    By focusing on the identification of animal behaviours through machine learning, we utilized features derived from accelerometer and gyroscope data to train decision trees of varying depths and combinations of sensor values and their encodings.
    We conducted an experiment to gain data which held respective sensor values and produced the varying decision trees.
    Furthermore, we investigated the impact of the different sensors and their encodings on some decision trees accuracy metrics.
    Our experimental results indicate that decision trees can effectively classify behaviours with high accuracy while maintaining reasonably high performance on-board.
    A key finding is the importance of gyroscope data, which consistently appeared in the top-performing feature combinations, particularly for distinguishing between behaviours such as standing and walking.
    This suggests that rotational data, as captured by the gyroscope, plays a critical role in behaviour recognition.
    Furthermore, we demonstrated that carefully selecting which data to transmit, based on the output of decision trees, can lead to substantial energy savings.
    By reducing the amount of data transmitted to only what is necessary for behaviour classification, we predicted significant reductions in energy consumption—up to $99.9903\%$ in some specific scenarios.
    This energy consumption reduction has profound implications for the operational longevity of bio-loggers, potentially extending their runtime from 94 to 137 days, as is the case for one example configuration of the WildFi tag.
    Thus, our findings bring together the feasibility and potential benefits of integrating machine learning models into bio-loggers for real-time, on-board data processing and energy-efficient operation.
    We show, that on-board classification on bio-loggers and the use of it as a data filter works, that it is feasible within reasonable limits and that it can significantly increase the runtime of a device.  
    
    In conclusion, this study contributes to the growing body of research on the intersection of animal behaviour monitoring and machine learning, offering insights into how even simple machine-learning approaches can be utilised to make bio-logging more energy-efficient, thereby enabling longer-term and more detailed wildlife studies.

\backmatter

\bmhead{Supplementary information}

\bmhead{Acknowledgements}
We acknowledge funding by DFG, under Germany’s Excellence Strategy – EXC 2117 – 422037984, and DFG project ID 251654672 – TRR 161.
Furthermore, the authors thank the Max-Planck-Institute of Animal Behaviour for their support.
We acknowledge the help of Nina Richter and Marcel Escher with the experiment.
\bmhead{Authors' contributions}
WKM was the main contributor to writing the manuscript and conducting the experiments.
FS and KK provided crucial input to theoretical concepts from the computer science domain.
TAW supported with answers to hardware-related issues, as well as domain knowledge from engineering.
All authors read and approved the final manuscript.
\bmhead{Funding}
We acknowledge funding by DFG, under Germany’s Excellence Strategy – EXC 2117 – 422037984, and DFG project ID 251654672 – TRR 161.
\bmhead{Availability of data and materials}
The datasets generated and analysed during the current study are available in the Zenodo repository, \url{https://doi.org/10.5281/zenodo.13643513}.

\section*{Declarations}

\subsection*{Ethics approval and consent to participate}
Not applicable.
\subsection*{Consent for publication}
Not applicable.
\subsection*{Competing interests}
The authors declare no competing interests.

\bibliography{sn-bibliography}


\begin{thebibliography}{50}
\ifx \bisbn   \undefined \def \bisbn  #1{ISBN #1}\fi
\ifx \binits  \undefined \def \binits#1{#1}\fi
\ifx \bauthor  \undefined \def \bauthor#1{#1}\fi
\ifx \batitle  \undefined \def \batitle#1{#1}\fi
\ifx \bjtitle  \undefined \def \bjtitle#1{#1}\fi
\ifx \bvolume  \undefined \def \bvolume#1{\textbf{#1}}\fi
\ifx \byear  \undefined \def \byear#1{#1}\fi
\ifx \bissue  \undefined \def \bissue#1{#1}\fi
\ifx \bfpage  \undefined \def \bfpage#1{#1}\fi
\ifx \blpage  \undefined \def \blpage #1{#1}\fi
\ifx \burl  \undefined \def \burl#1{\textsf{#1}}\fi
\ifx \doiurl  \undefined \def \doiurl#1{\url{https://doi.org/#1}}\fi
\ifx \betal  \undefined \def \betal{\textit{et al.}}\fi
\ifx \binstitute  \undefined \def \binstitute#1{#1}\fi
\ifx \binstitutionaled  \undefined \def \binstitutionaled#1{#1}\fi
\ifx \bctitle  \undefined \def \bctitle#1{#1}\fi
\ifx \beditor  \undefined \def \beditor#1{#1}\fi
\ifx \bpublisher  \undefined \def \bpublisher#1{#1}\fi
\ifx \bbtitle  \undefined \def \bbtitle#1{#1}\fi
\ifx \bedition  \undefined \def \bedition#1{#1}\fi
\ifx \bseriesno  \undefined \def \bseriesno#1{#1}\fi
\ifx \blocation  \undefined \def \blocation#1{#1}\fi
\ifx \bsertitle  \undefined \def \bsertitle#1{#1}\fi
\ifx \bsnm \undefined \def \bsnm#1{#1}\fi
\ifx \bsuffix \undefined \def \bsuffix#1{#1}\fi
\ifx \bparticle \undefined \def \bparticle#1{#1}\fi
\ifx \barticle \undefined \def \barticle#1{#1}\fi
\bibcommenthead
\ifx \bconfdate \undefined \def \bconfdate #1{#1}\fi
\ifx \botherref \undefined \def \botherref #1{#1}\fi
\ifx \url \undefined \def \url#1{\textsf{#1}}\fi
\ifx \bchapter \undefined \def \bchapter#1{#1}\fi
\ifx \bbook \undefined \def \bbook#1{#1}\fi
\ifx \bcomment \undefined \def \bcomment#1{#1}\fi
\ifx \oauthor \undefined \def \oauthor#1{#1}\fi
\ifx \citeauthoryear \undefined \def \citeauthoryear#1{#1}\fi
\ifx \endbibitem  \undefined \def \endbibitem {}\fi
\ifx \bconflocation  \undefined \def \bconflocation#1{#1}\fi
\ifx \arxivurl  \undefined \def \arxivurl#1{\textsf{#1}}\fi
\csname PreBibitemsHook\endcsname

\bibitem[\protect\citeauthoryear{Richter}{1927}]{richter1927animal}
\begin{barticle}
\bauthor{\bsnm{Richter}, \binits{C.P.}}:
\batitle{Animal behavior and internal drives}.
\bjtitle{The Quarterly Review of Biology}
\bvolume{2}(\bissue{3}),
\bfpage{307}--\blpage{343}
(\byear{1927})
\end{barticle}
\endbibitem

\bibitem[\protect\citeauthoryear{Beach and Jaynes}{1954}]{beach1954effects}
\begin{barticle}
\bauthor{\bsnm{Beach}, \binits{F.A.}},
\bauthor{\bsnm{Jaynes}, \binits{J.}}:
\batitle{Effects of early experience upon the behavior of animals.}
\bjtitle{Psychological Bulletin}
\bvolume{51}(\bissue{3}),
\bfpage{239}
(\byear{1954})
\end{barticle}
\endbibitem

\bibitem[\protect\citeauthoryear{McFarland}{1971}]{mcfarland1971feedback}
\begin{bbook}
\bauthor{\bsnm{McFarland}, \binits{D.J.}}:
\bbtitle{Feedback Mechanisms in Animal Behaviour.},
(\byear{1971})
\end{bbook}
\endbibitem

\bibitem[\protect\citeauthoryear{Manning and Dawkins}{2012}]{manning2012introduction}
\begin{bbook}
\bauthor{\bsnm{Manning}, \binits{A.}},
\bauthor{\bsnm{Dawkins}, \binits{M.S.}}:
\bbtitle{An Introduction to Animal Behaviour},
pp. \bfpage{5}--\blpage{7}
(\byear{2012})
\end{bbook}
\endbibitem

\bibitem[\protect\citeauthoryear{Kooyman}{1965}]{Kooyman1965}
\begin{barticle}
\bauthor{\bsnm{Kooyman}, \binits{G.L.}}:
\batitle{Techniques used in measuring diving capacities of weddell seals}.
\bjtitle{Polar Record}
\bvolume{12}(\bissue{79}),
\bfpage{391}--\blpage{394}
(\byear{1965})
\end{barticle}
\endbibitem

\bibitem[\protect\citeauthoryear{Guilford et~al.}{2008}]{GUILFORD2008}
\begin{barticle}
\bauthor{\bsnm{Guilford}, \binits{T.C.}},
\bauthor{\bsnm{Meade}, \binits{J.}},
\bauthor{\bsnm{Freeman}, \binits{R.}},
\bauthor{\bsnm{Biro}, \binits{D.}},
\bauthor{\bsnm{Evans}, \binits{T.}},
\bauthor{\bsnm{Bonadonna}, \binits{F.}},
\bauthor{\bsnm{Doyle}, \binits{D.}},
\bauthor{\bsnm{Roberts}, \binits{S.}},
\bauthor{\bsnm{Perrins}, \binits{C.M.}}:
\batitle{Gps tracking of the foraging movements of manx shearwaters puffinus puffinus breeding on skomer island, wales}.
\bjtitle{Ibis}
\bvolume{150}(\bissue{3}),
\bfpage{462}--\blpage{473}
(\byear{2008})
\end{barticle}
\endbibitem

\bibitem[\protect\citeauthoryear{Kojola et~al.}{2006}]{kojola2006dispersal}
\begin{barticle}
\bauthor{\bsnm{Kojola}, \binits{I.}},
\bauthor{\bsnm{Aspi}, \binits{J.}},
\bauthor{\bsnm{Hakala}, \binits{A.}},
\bauthor{\bsnm{Heikkinen}, \binits{S.}},
\bauthor{\bsnm{Ilmoni}, \binits{C.}},
\bauthor{\bsnm{Ronkainen}, \binits{S.}}:
\batitle{Dispersal in an expanding wolf population in finland}.
\bjtitle{Journal of Mammalogy}
\bvolume{87}(\bissue{2}),
\bfpage{281}--\blpage{286}
(\byear{2006})
\end{barticle}
\endbibitem

\bibitem[\protect\citeauthoryear{Sand et~al.}{2005}]{sand2005using}
\begin{barticle}
\bauthor{\bsnm{Sand}, \binits{H.}},
\bauthor{\bsnm{Zimmermann}, \binits{B.}},
\bauthor{\bsnm{Wabakken}, \binits{P.}},
\bauthor{\bsnm{Andr{\`e}n}, \binits{H.}},
\bauthor{\bsnm{Pedersen}, \binits{H.C.}}:
\batitle{Using gps technology and gis cluster analyses to estimate kill rates in wolf-ungulate ecosystems}.
\bjtitle{Wildlife Society Bulletin}
\bvolume{33}(\bissue{3}),
\bfpage{914}--\blpage{925}
(\byear{2005})
\end{barticle}
\endbibitem

\bibitem[\protect\citeauthoryear{Brown et~al.}{2013}]{brown2013observing}
\begin{barticle}
\bauthor{\bsnm{Brown}, \binits{D.D.}},
\bauthor{\bsnm{Kays}, \binits{R.}},
\bauthor{\bsnm{Wikelski}, \binits{M.}},
\bauthor{\bsnm{Wilson}, \binits{R.}},
\bauthor{\bsnm{Klimley}, \binits{A.P.}}:
\batitle{Observing the unwatchable through acceleration logging of animal behavior}.
\bjtitle{Animal Biotelemetry}
\bvolume{1},
\bfpage{1}--\blpage{16}
(\byear{2013})
\end{barticle}
\endbibitem

\bibitem[\protect\citeauthoryear{Gleiss et~al.}{2010}]{GLEISS201085}
\begin{barticle}
\bauthor{\bsnm{Gleiss}, \binits{A.C.}},
\bauthor{\bsnm{Dale}, \binits{J.J.}},
\bauthor{\bsnm{Holland}, \binits{K.N.}},
\bauthor{\bsnm{Wilson}, \binits{R.P.}}:
\batitle{Accelerating estimates of activity-specific metabolic rate in fishes: Testing the applicability of acceleration data-loggers}.
\bjtitle{Journal of Experimental Marine Biology and Ecology}
\bvolume{385}(\bissue{1}),
\bfpage{85}--\blpage{91}
(\byear{2010})
\end{barticle}
\endbibitem

\bibitem[\protect\citeauthoryear{Kr{\"o}schel et~al.}{2017}]{kroschel2017remote}
\begin{barticle}
\bauthor{\bsnm{Kr{\"o}schel}, \binits{M.}},
\bauthor{\bsnm{Reineking}, \binits{B.}},
\bauthor{\bsnm{Werwie}, \binits{F.}},
\bauthor{\bsnm{Wildi}, \binits{F.}},
\bauthor{\bsnm{Storch}, \binits{I.}}:
\batitle{Remote monitoring of vigilance behavior in large herbivores using acceleration data}.
\bjtitle{Animal Biotelemetry}
\bvolume{5},
\bfpage{1}--\blpage{15}
(\byear{2017})
\end{barticle}
\endbibitem

\bibitem[\protect\citeauthoryear{Krone et~al.}{2008}]{Krone2008}
\begin{barticle}
\bauthor{\bsnm{Krone}, \binits{O.}},
\bauthor{\bsnm{Berger}, \binits{A.}},
\bauthor{\bsnm{Schulte}, \binits{R.}}:
\batitle{Recording movement and activity pattern of a white-tailed sea eagle (haliaeetus albicilla) by a {GPS} datalogger}.
\bjtitle{Journal of Ornithology}
\bvolume{150}(\bissue{1}),
\bfpage{273}--\blpage{280}
(\byear{2008})
\end{barticle}
\endbibitem

\bibitem[\protect\citeauthoryear{Hooker et~al.}{2007}]{hooker2007bio}
\begin{barticle}
\bauthor{\bsnm{Hooker}, \binits{S.K.}},
\bauthor{\bsnm{Biuw}, \binits{M.}},
\bauthor{\bsnm{McConnell}, \binits{B.J.}},
\bauthor{\bsnm{Miller}, \binits{P.J.}},
\bauthor{\bsnm{Sparling}, \binits{C.E.}}:
\batitle{Bio-logging science: logging and relaying physical and biological data using animal-attached tags}.
\bjtitle{Deep-Sea Research Part II}
\bvolume{3}(\bissue{54}),
\bfpage{177}--\blpage{182}
(\byear{2007})
\end{barticle}
\endbibitem

\bibitem[\protect\citeauthoryear{Andrzejaczek et~al.}{2019}]{andrzejaczek2019biologging}
\begin{barticle}
\bauthor{\bsnm{Andrzejaczek}, \binits{S.}},
\bauthor{\bsnm{Gleiss}, \binits{A.C.}},
\bauthor{\bsnm{Lear}, \binits{K.O.}},
\bauthor{\bsnm{Pattiaratchi}, \binits{C.B.}},
\bauthor{\bsnm{Chapple}, \binits{T.K.}},
\bauthor{\bsnm{Meekan}, \binits{M.G.}}:
\batitle{Biologging tags reveal links between fine-scale horizontal and vertical movement behaviors in tiger sharks (galeocerdo cuvier)}.
\bjtitle{Frontiers in Marine Science}
\bvolume{6},
\bfpage{229}
(\byear{2019})
\end{barticle}
\endbibitem

\bibitem[\protect\citeauthoryear{Fourati et~al.}{2011}]{Fourati2011}
\begin{barticle}
\bauthor{\bsnm{Fourati}, \binits{H.}},
\bauthor{\bsnm{Manamanni}, \binits{N.}},
\bauthor{\bsnm{Afilal}, \binits{L.}},
\bauthor{\bsnm{Handrich}, \binits{Y.}}:
\batitle{Posture and body acceleration tracking by inertial and magnetic sensing: Application in behavioral analysis of free-ranging animals}.
\bjtitle{Biomedical Signal Processing and Control}
\bvolume{6}(\bissue{1}),
\bfpage{94}--\blpage{104}
(\byear{2011})
\end{barticle}
\endbibitem

\bibitem[\protect\citeauthoryear{Ware et~al.}{2016}]{ware2016averaged}
\begin{barticle}
\bauthor{\bsnm{Ware}, \binits{C.}},
\bauthor{\bsnm{Trites}, \binits{A.W.}},
\bauthor{\bsnm{Rosen}, \binits{D.A.}},
\bauthor{\bsnm{Potvin}, \binits{J.}}:
\batitle{Averaged propulsive body acceleration (apba) can be calculated from biologging tags that incorporate gyroscopes and accelerometers to estimate swimming speed, hydrodynamic drag and energy expenditure for steller sea lions}.
\bjtitle{PloS one}
\bvolume{11}(\bissue{6}),
\bfpage{0157326}
(\byear{2016})
\end{barticle}
\endbibitem

\bibitem[\protect\citeauthoryear{Chakravarty et~al.}{2019}]{chakravarty2019novel}
\begin{barticle}
\bauthor{\bsnm{Chakravarty}, \binits{P.}},
\bauthor{\bsnm{Cozzi}, \binits{G.}},
\bauthor{\bsnm{Ozgul}, \binits{A.}},
\bauthor{\bsnm{Aminian}, \binits{K.}}:
\batitle{A novel biomechanical approach for animal behaviour recognition using accelerometers}.
\bjtitle{Methods in Ecology and Evolution}
\bvolume{10}(\bissue{6}),
\bfpage{802}--\blpage{814}
(\byear{2019})
\end{barticle}
\endbibitem

\bibitem[\protect\citeauthoryear{Chakravarty et~al.}{2020}]{chakravarty2020seek}
\begin{barticle}
\bauthor{\bsnm{Chakravarty}, \binits{P.}},
\bauthor{\bsnm{Cozzi}, \binits{G.}},
\bauthor{\bsnm{Dejnabadi}, \binits{H.}},
\bauthor{\bsnm{L{\'e}ziart}, \binits{P.-A.}},
\bauthor{\bsnm{Manser}, \binits{M.}},
\bauthor{\bsnm{Ozgul}, \binits{A.}},
\bauthor{\bsnm{Aminian}, \binits{K.}}:
\batitle{Seek and learn: Automated identification of microevents in animal behaviour using envelopes of acceleration data and machine learning}.
\bjtitle{Methods in Ecology and Evolution}
\bvolume{11}(\bissue{12}),
\bfpage{1639}--\blpage{1651}
(\byear{2020})
\end{barticle}
\endbibitem

\bibitem[\protect\citeauthoryear{Gauld et~al.}{2023}]{Gauld2023}
\begin{botherref}
\oauthor{\bsnm{Gauld}, \binits{J.}},
\oauthor{\bsnm{Atkinson}, \binits{P.W.}},
\oauthor{\bsnm{Silva}, \binits{J.P.}},
\oauthor{\bsnm{Senn}, \binits{A.}},
\oauthor{\bsnm{Franco}, \binits{A.M.A.}}:
Characterisation of a new lightweight {LoRaWAN} {GPS} bio-logger and deployment on griffon vultures gyps fulvus.
Animal Biotelemetry
\textbf{11}(1)
(2023)
\end{botherref}
\endbibitem

\bibitem[\protect\citeauthoryear{Wild et~al.}{2022}]{Wild2022}
\begin{barticle}
\bauthor{\bsnm{Wild}, \binits{T.A.}},
\bauthor{\bsnm{Wikelski}, \binits{M.}},
\bauthor{\bsnm{Tyndel}, \binits{S.}},
\bauthor{\bsnm{Alarc{\'{o}}n-Nieto}, \binits{G.}},
\bauthor{\bsnm{Klump}, \binits{B.C.}},
\bauthor{\bsnm{Aplin}, \binits{L.M.}},
\bauthor{\bsnm{Meboldt}, \binits{M.}},
\bauthor{\bsnm{Williams}, \binits{H.J.}}:
\batitle{Internet on animals: Wi-fi-enabled devices provide a solution for big data transmission in biologging}.
\bjtitle{Methods in Ecology and Evolution}
\bvolume{14}(\bissue{1}),
\bfpage{87}--\blpage{102}
(\byear{2022})
\end{barticle}
\endbibitem

\bibitem[\protect\citeauthoryear{Ayele et~al.}{2018}]{ayele2018towards}
\begin{bchapter}
\bauthor{\bsnm{Ayele}, \binits{E.D.}},
\bauthor{\bsnm{Meratnia}, \binits{N.}},
\bauthor{\bsnm{Havinga}, \binits{P.J.}}:
\bctitle{Towards a new opportunistic iot network architecture for wildlife monitoring system}.
In: \bbtitle{2018 9th IFIP International Conference on New Technologies, Mobility and Security (NTMS)},
pp. \bfpage{1}--\blpage{5}
(\byear{2018})
\end{bchapter}
\endbibitem

\bibitem[\protect\citeauthoryear{Krondorf et~al.}{2022}]{krondorf2022icarus}
\begin{barticle}
\bauthor{\bsnm{Krondorf}, \binits{M.}},
\bauthor{\bsnm{Bittner}, \binits{S.}},
\bauthor{\bsnm{Plettemeier}, \binits{D.}},
\bauthor{\bsnm{Knopp}, \binits{A.}},
\bauthor{\bsnm{Wikelski}, \binits{M.}}:
\batitle{Icarus—very low power satellite-based iot}.
\bjtitle{Sensors}
\bvolume{22}(\bissue{17}),
\bfpage{6329}
(\byear{2022})
\end{barticle}
\endbibitem

\bibitem[\protect\citeauthoryear{Bridge et~al.}{2011}]{Bridge2011}
\begin{barticle}
\bauthor{\bsnm{Bridge}, \binits{E.S.}},
\bauthor{\bsnm{Thorup}, \binits{K.}},
\bauthor{\bsnm{Bowlin}, \binits{M.S.}},
\bauthor{\bsnm{Chilson}, \binits{P.B.}},
\bauthor{\bsnm{Diehl}, \binits{R.H.}},
\bauthor{\bsnm{Fléron}, \binits{R.W.}},
\bauthor{\bsnm{Hartl}, \binits{P.}},
\bauthor{\bsnm{Kays}, \binits{R.}},
\bauthor{\bsnm{Kelly}, \binits{J.F.}},
\bauthor{\bsnm{Robinson}, \binits{W.D.}},
\bauthor{\bsnm{Wikelski}, \binits{M.}}:
\batitle{Technology on the move: Recent and forthcoming innovations for tracking migratory birds}.
\bjtitle{BioScience}
\bvolume{61}(\bissue{9}),
\bfpage{689}--\blpage{698}
(\byear{2011})
\end{barticle}
\endbibitem

\bibitem[\protect\citeauthoryear{Kays et~al.}{2015}]{Kays2015}
\begin{botherref}
\oauthor{\bsnm{Kays}, \binits{R.}},
\oauthor{\bsnm{Crofoot}, \binits{M.C.}},
\oauthor{\bsnm{Jetz}, \binits{W.}},
\oauthor{\bsnm{Wikelski}, \binits{M.}}:
Terrestrial animal tracking as an eye on life and planet.
Science
\textbf{348}(6240)
(2015)
\end{botherref}
\endbibitem

\bibitem[\protect\citeauthoryear{Wild et~al.}{2023}]{wild2023multi}
\begin{barticle}
\bauthor{\bsnm{Wild}, \binits{T.A.}},
\bauthor{\bsnm{Schalkwyk}, \binits{L.}},
\bauthor{\bsnm{Viljoen}, \binits{P.}},
\bauthor{\bsnm{Heine}, \binits{G.}},
\bauthor{\bsnm{Richter}, \binits{N.}},
\bauthor{\bsnm{Vorneweg}, \binits{B.}},
\bauthor{\bsnm{Koblitz}, \binits{J.C.}},
\bauthor{\bsnm{Dechmann}, \binits{D.K.}},
\bauthor{\bsnm{Rogers}, \binits{W.}},
\bauthor{\bsnm{Partecke}, \binits{J.}}, \betal:
\batitle{A multi-species evaluation of digital wildlife monitoring using the sigfox iot network}.
\bjtitle{Animal Biotelemetry}
\bvolume{11}(\bissue{1}),
\bfpage{13}
(\byear{2023})
\end{barticle}
\endbibitem

\bibitem[\protect\citeauthoryear{Azar et~al.}{2019}]{AZAR2019168}
\begin{barticle}
\bauthor{\bsnm{Azar}, \binits{J.}},
\bauthor{\bsnm{Makhoul}, \binits{A.}},
\bauthor{\bsnm{Barhamgi}, \binits{M.}},
\bauthor{\bsnm{Couturier}, \binits{R.}}:
\batitle{An energy efficient iot data compression approach for edge machine learning}.
\bjtitle{Future Generation Computer Systems}
\bvolume{96},
\bfpage{168}--\blpage{175}
(\byear{2019})
\end{barticle}
\endbibitem

\bibitem[\protect\citeauthoryear{Humayun et~al.}{2022}]{humayun2022energy}
\begin{barticle}
\bauthor{\bsnm{Humayun}, \binits{M.}},
\bauthor{\bsnm{Alsaqer}, \binits{M.S.}},
\bauthor{\bsnm{Jhanjhi}, \binits{N.}}:
\batitle{Energy optimization for smart cities using iot}.
\bjtitle{Applied Artificial Intelligence}
\bvolume{36}(\bissue{1}),
\bfpage{2037255}
(\byear{2022})
\end{barticle}
\endbibitem

\bibitem[\protect\citeauthoryear{Di{\`e}ne et~al.}{2020}]{diene2020data}
\begin{barticle}
\bauthor{\bsnm{Di{\`e}ne}, \binits{B.}},
\bauthor{\bsnm{Rodrigues}, \binits{J.J.}},
\bauthor{\bsnm{Diallo}, \binits{O.}},
\bauthor{\bsnm{Ndoye}, \binits{E.H.M.}},
\bauthor{\bsnm{Korotaev}, \binits{V.V.}}:
\batitle{Data management techniques for internet of things}.
\bjtitle{Mechanical Systems and Signal Processing}
\bvolume{138},
\bfpage{106564}
(\byear{2020})
\end{barticle}
\endbibitem

\bibitem[\protect\citeauthoryear{Saqlain et~al.}{2019}]{Saqlain2019}
\begin{barticle}
\bauthor{\bsnm{Saqlain}},
\bauthor{\bsnm{Piao}},
\bauthor{\bsnm{Shim}},
\bauthor{\bsnm{Lee}}:
\batitle{Framework of an iot-based industrial data management for smart manufacturing}.
\bjtitle{Journal of Sensor and Actuator Networks}
\bvolume{8}(\bissue{2}),
\bfpage{25}
(\byear{2019})
\end{barticle}
\endbibitem

\bibitem[\protect\citeauthoryear{Salah et~al.}{2020}]{salah2020iot}
\begin{bchapter}
\bauthor{\bsnm{Salah}, \binits{K.}},
\bauthor{\bsnm{Alfalasi}, \binits{A.}},
\bauthor{\bsnm{Alfalasi}, \binits{M.}},
\bauthor{\bsnm{Alharmoudi}, \binits{M.}},
\bauthor{\bsnm{Alzaabi}, \binits{M.}},
\bauthor{\bsnm{Alzyeodi}, \binits{A.}},
\bauthor{\bsnm{Ahmad}, \binits{R.W.}}:
\bctitle{Iot-enabled shipping container with environmental monitoring and location tracking}.
In: \bbtitle{2020 IEEE 17th Annual Consumer Communications \& Networking Conference (CCNC)},
pp. \bfpage{1}--\blpage{6}
(\byear{2020})
\end{bchapter}
\endbibitem

\bibitem[\protect\citeauthoryear{Bidder et~al.}{2020}]{Bidder2020}
\begin{botherref}
\oauthor{\bsnm{Bidder}, \binits{O.R.}},
\oauthor{\bsnm{Virgilio}, \binits{A.}},
\oauthor{\bsnm{Hunter}, \binits{J.S.}},
\oauthor{\bsnm{McInturff}, \binits{A.}},
\oauthor{\bsnm{Gaynor}, \binits{K.M.}},
\oauthor{\bsnm{Smith}, \binits{A.M.}},
\oauthor{\bsnm{Dorcy}, \binits{J.}},
\oauthor{\bsnm{Rosell}, \binits{F.}}:
Monitoring canid scent marking in space and time using a biologging and machine learning approach.
Scientific Reports
\textbf{10}(1)
(2020)
\end{botherref}
\endbibitem

\bibitem[\protect\citeauthoryear{Brandes et~al.}{2021}]{brandes2021behaviour}
\begin{barticle}
\bauthor{\bsnm{Brandes}, \binits{S.}},
\bauthor{\bsnm{Sicks}, \binits{F.}},
\bauthor{\bsnm{Berger}, \binits{A.}}:
\batitle{Behaviour classification on giraffes (giraffa camelopardalis) using machine learning algorithms on triaxial acceleration data of two commonly used gps devices and its possible application for their management and conservation}.
\bjtitle{Sensors}
\bvolume{21}(\bissue{6}),
\bfpage{2229}
(\byear{2021})
\end{barticle}
\endbibitem

\bibitem[\protect\citeauthoryear{Tuia et~al.}{2022}]{tuia2022perspectives}
\begin{barticle}
\bauthor{\bsnm{Tuia}, \binits{D.}},
\bauthor{\bsnm{Kellenberger}, \binits{B.}},
\bauthor{\bsnm{Beery}, \binits{S.}},
\bauthor{\bsnm{Costelloe}, \binits{B.R.}},
\bauthor{\bsnm{Zuffi}, \binits{S.}},
\bauthor{\bsnm{Risse}, \binits{B.}},
\bauthor{\bsnm{Mathis}, \binits{A.}},
\bauthor{\bsnm{Mathis}, \binits{M.W.}},
\bauthor{\bsnm{Van~Langevelde}, \binits{F.}},
\bauthor{\bsnm{Burghardt}, \binits{T.}}, \betal:
\batitle{Perspectives in machine learning for wildlife conservation}.
\bjtitle{Nature communications}
\bvolume{13}(\bissue{1}),
\bfpage{1}--\blpage{15}
(\byear{2022})
\end{barticle}
\endbibitem

\bibitem[\protect\citeauthoryear{Bidder et~al.}{2014}]{bidder2014love}
\begin{barticle}
\bauthor{\bsnm{Bidder}, \binits{O.R.}},
\bauthor{\bsnm{Campbell}, \binits{H.A.}},
\bauthor{\bsnm{G{\'o}mez-Laich}, \binits{A.}},
\bauthor{\bsnm{Urg{\'e}}, \binits{P.}},
\bauthor{\bsnm{Walker}, \binits{J.}},
\bauthor{\bsnm{Cai}, \binits{Y.}},
\bauthor{\bsnm{Gao}, \binits{L.}},
\bauthor{\bsnm{Quintana}, \binits{F.}},
\bauthor{\bsnm{Wilson}, \binits{R.P.}}:
\batitle{Love thy neighbour: automatic animal behavioural classification of acceleration data using the k-nearest neighbour algorithm}.
\bjtitle{PloS one}
\bvolume{9}(\bissue{2}),
\bfpage{88609}
(\byear{2014})
\end{barticle}
\endbibitem

\bibitem[\protect\citeauthoryear{Kadar et~al.}{2020}]{kadar2020assessment}
\begin{barticle}
\bauthor{\bsnm{Kadar}, \binits{J.P.}},
\bauthor{\bsnm{Ladds}, \binits{M.A.}},
\bauthor{\bsnm{Day}, \binits{J.}},
\bauthor{\bsnm{Lyall}, \binits{B.}},
\bauthor{\bsnm{Brown}, \binits{C.}}:
\batitle{Assessment of machine learning models to identify port jackson shark behaviours using tri-axial accelerometers}.
\bjtitle{Sensors}
\bvolume{20}(\bissue{24}),
\bfpage{7096}
(\byear{2020})
\end{barticle}
\endbibitem

\bibitem[\protect\citeauthoryear{Wang}{2019}]{wang2019machine}
\begin{barticle}
\bauthor{\bsnm{Wang}, \binits{G.}}:
\batitle{Machine learning for inferring animal behavior from location and movement data}.
\bjtitle{Ecological informatics}
\bvolume{49},
\bfpage{69}--\blpage{76}
(\byear{2019})
\end{barticle}
\endbibitem

\bibitem[\protect\citeauthoryear{Korpela et~al.}{2020}]{Korpela2020}
\begin{botherref}
\oauthor{\bsnm{Korpela}, \binits{J.}},
\oauthor{\bsnm{Suzuki}, \binits{H.}},
\oauthor{\bsnm{Matsumoto}, \binits{S.}},
\oauthor{\bsnm{Mizutani}, \binits{Y.}},
\oauthor{\bsnm{Samejima}, \binits{M.}},
\oauthor{\bsnm{Maekawa}, \binits{T.}},
\oauthor{\bsnm{Nakai}, \binits{J.}},
\oauthor{\bsnm{Yoda}, \binits{K.}}:
Machine learning enables improved runtime and precision for bio-loggers on seabirds.
Communications Biology
\textbf{3}(1)
(2020)
\end{botherref}
\endbibitem

\bibitem[\protect\citeauthoryear{{Bosch}}{}]{bmx160}
\begin{botherref}
\oauthor{\bsnm{{Bosch}}}:
BMX160.
\url{https://www.mouser.com/pdfdocs/BST-BMX160-DS000-11.pdf}
\end{botherref}
\endbibitem

\bibitem[\protect\citeauthoryear{{Bosch}}{}]{bme680}
\begin{botherref}
\oauthor{\bsnm{{Bosch}}}:
BME680.
\url{https://www.bosch-sensortec.com/media/boschsensortec/downloads/datasheets/bst-bme680-ds001.pdf}
\end{botherref}
\endbibitem

\bibitem[\protect\citeauthoryear{Pedregosa et~al.}{2011}]{pedregosa2011scikit}
\begin{barticle}
\bauthor{\bsnm{Pedregosa}, \binits{F.}},
\bauthor{\bsnm{Varoquaux}, \binits{G.}},
\bauthor{\bsnm{Gramfort}, \binits{A.}},
\bauthor{\bsnm{Michel}, \binits{V.}},
\bauthor{\bsnm{Thirion}, \binits{B.}},
\bauthor{\bsnm{Grisel}, \binits{O.}},
\bauthor{\bsnm{Blondel}, \binits{M.}},
\bauthor{\bsnm{Prettenhofer}, \binits{P.}},
\bauthor{\bsnm{Weiss}, \binits{R.}},
\bauthor{\bsnm{Dubourg}, \binits{V.}}, \betal:
\batitle{Scikit-learn: Machine learning in python}.
\bjtitle{the Journal of machine Learning research}
\bvolume{12},
\bfpage{2825}--\blpage{2830}
(\byear{2011})
\end{barticle}
\endbibitem

\bibitem[\protect\citeauthoryear{Jain et~al.}{2018}]{jain2018gist}
\begin{bchapter}
\bauthor{\bsnm{Jain}, \binits{A.}},
\bauthor{\bsnm{Phanishayee}, \binits{A.}},
\bauthor{\bsnm{Mars}, \binits{J.}},
\bauthor{\bsnm{Tang}, \binits{L.}},
\bauthor{\bsnm{Pekhimenko}, \binits{G.}}:
\bctitle{Gist: Efficient data encoding for deep neural network training}.
In: \bbtitle{2018 ACM/IEEE 45th Annual International Symposium on Computer Architecture (ISCA)},
pp. \bfpage{776}--\blpage{789}
(\byear{2018})
\end{bchapter}
\endbibitem

\bibitem[\protect\citeauthoryear{Lelewer and Hirschberg}{1987}]{lelewer1987data}
\begin{barticle}
\bauthor{\bsnm{Lelewer}, \binits{D.A.}},
\bauthor{\bsnm{Hirschberg}, \binits{D.S.}}:
\batitle{Data compression}.
\bjtitle{ACM Computing Surveys (CSUR)}
\bvolume{19}(\bissue{3}),
\bfpage{261}--\blpage{296}
(\byear{1987})
\end{barticle}
\endbibitem

\bibitem[\protect\citeauthoryear{Gleiss et~al.}{2011}]{gleiss2011}
\begin{barticle}
\bauthor{\bsnm{Gleiss}, \binits{A.C.}},
\bauthor{\bsnm{Wilson}, \binits{R.P.}},
\bauthor{\bsnm{Shepard}, \binits{E.L.C.}}:
\batitle{Making overall dynamic body acceleration work: on the theory of acceleration as a proxy for energy expenditure}.
\bjtitle{Methods in Ecology and Evolution}
\bvolume{2}(\bissue{1}),
\bfpage{23}--\blpage{33}
(\byear{2011})
\end{barticle}
\endbibitem

\bibitem[\protect\citeauthoryear{L{\'{o}}pez et~al.}{2021}]{Lopez2021}
\begin{barticle}
\bauthor{\bsnm{L{\'{o}}pez}, \binits{L.M.M.}},
\bauthor{\bsnm{Soto}, \binits{N.A.}},
\bauthor{\bsnm{Madsen}, \binits{P.T.}},
\bauthor{\bsnm{Johnson}, \binits{M.}}:
\batitle{Overall dynamic body acceleration measures activity differently on large versus small aquatic animals}.
\bjtitle{Methods in Ecology and Evolution}
\bvolume{13}(\bissue{2}),
\bfpage{447}--\blpage{458}
(\byear{2021})
\end{barticle}
\endbibitem

\bibitem[\protect\citeauthoryear{Bulling et~al.}{2014}]{bulling2014tutorial}
\begin{barticle}
\bauthor{\bsnm{Bulling}, \binits{A.}},
\bauthor{\bsnm{Blanke}, \binits{U.}},
\bauthor{\bsnm{Schiele}, \binits{B.}}:
\batitle{A tutorial on human activity recognition using body-worn inertial sensors}.
\bjtitle{ACM Computing Surveys (CSUR)}
\bvolume{46}(\bissue{3}),
\bfpage{1}--\blpage{33}
(\byear{2014})
\end{barticle}
\endbibitem

\bibitem[\protect\citeauthoryear{Williams et~al.}{2019}]{Williams2019}
\begin{barticle}
\bauthor{\bsnm{Williams}, \binits{H.J.}},
\bauthor{\bsnm{Taylor}, \binits{L.A.}},
\bauthor{\bsnm{Benhamou}, \binits{S.}},
\bauthor{\bsnm{Bijleveld}, \binits{A.I.}},
\bauthor{\bsnm{Clay}, \binits{T.A.}},
\bauthor{\bsnm{Grissac}, \binits{S.}},
\bauthor{\bsnm{Dem{\v{s}}ar}, \binits{U.}},
\bauthor{\bsnm{English}, \binits{H.M.}},
\bauthor{\bsnm{Franconi}, \binits{N.}},
\bauthor{\bsnm{G{\'{o}}mez-Laich}, \binits{A.}},
\bauthor{\bsnm{Griffiths}, \binits{R.C.}},
\bauthor{\bsnm{Kay}, \binits{W.P.}},
\bauthor{\bsnm{Morales}, \binits{J.M.}},
\bauthor{\bsnm{Potts}, \binits{J.R.}},
\bauthor{\bsnm{Rogerson}, \binits{K.F.}},
\bauthor{\bsnm{Rutz}, \binits{C.}},
\bauthor{\bsnm{Spelt}, \binits{A.}},
\bauthor{\bsnm{Trevail}, \binits{A.M.}},
\bauthor{\bsnm{Wilson}, \binits{R.P.}},
\bauthor{\bsnm{Börger}, \binits{L.}}:
\batitle{Optimizing the use of biologgers for movement ecology research}.
\bjtitle{Journal of Animal Ecology}
\bvolume{89}(\bissue{1}),
\bfpage{186}--\blpage{206}
(\byear{2019})
\end{barticle}
\endbibitem

\bibitem[\protect\citeauthoryear{Dutta and Bharali}{2021}]{dutta2021tinyml}
\begin{barticle}
\bauthor{\bsnm{Dutta}, \binits{L.}},
\bauthor{\bsnm{Bharali}, \binits{S.}}:
\batitle{Tinyml meets iot: A comprehensive survey}.
\bjtitle{Internet of Things}
\bvolume{16},
\bfpage{100461}
(\byear{2021})
\end{barticle}
\endbibitem

\bibitem[\protect\citeauthoryear{Abadi et~al.}{2015}]{tensorflow2015-whitepaper}
\begin{botherref}
\oauthor{\bsnm{Abadi}, \binits{M.}},
\oauthor{\bsnm{Agarwal}, \binits{A.}},
\oauthor{\bsnm{Barham}, \binits{P.}},
\oauthor{\bsnm{Brevdo}, \binits{E.}},
\oauthor{\bsnm{Chen}, \binits{Z.}},
\oauthor{\bsnm{Citro}, \binits{C.}},
\oauthor{\bsnm{Corrado}, \binits{G.S.}},
\oauthor{\bsnm{Davis}, \binits{A.}},
\oauthor{\bsnm{Dean}, \binits{J.}},
\oauthor{\bsnm{Devin}, \binits{M.}},
\oauthor{\bsnm{Ghemawat}, \binits{S.}},
\oauthor{\bsnm{Goodfellow}, \binits{I.}},
\oauthor{\bsnm{Harp}, \binits{A.}},
\oauthor{\bsnm{Irving}, \binits{G.}},
\oauthor{\bsnm{Isard}, \binits{M.}},
\oauthor{\bsnm{Jia}, \binits{Y.}},
\oauthor{\bsnm{Jozefowicz}, \binits{R.}},
\oauthor{\bsnm{Kaiser}, \binits{L.}},
\oauthor{\bsnm{Kudlur}, \binits{M.}},
\oauthor{\bsnm{Levenberg}, \binits{J.}},
\oauthor{\bsnm{Man\'{e}}, \binits{D.}},
\oauthor{\bsnm{Monga}, \binits{R.}},
\oauthor{\bsnm{Moore}, \binits{S.}},
\oauthor{\bsnm{Murray}, \binits{D.}},
\oauthor{\bsnm{Olah}, \binits{C.}},
\oauthor{\bsnm{Schuster}, \binits{M.}},
\oauthor{\bsnm{Shlens}, \binits{J.}},
\oauthor{\bsnm{Steiner}, \binits{B.}},
\oauthor{\bsnm{Sutskever}, \binits{I.}},
\oauthor{\bsnm{Talwar}, \binits{K.}},
\oauthor{\bsnm{Tucker}, \binits{P.}},
\oauthor{\bsnm{Vanhoucke}, \binits{V.}},
\oauthor{\bsnm{Vasudevan}, \binits{V.}},
\oauthor{\bsnm{Vi\'{e}gas}, \binits{F.}},
\oauthor{\bsnm{Vinyals}, \binits{O.}},
\oauthor{\bsnm{Warden}, \binits{P.}},
\oauthor{\bsnm{Wattenberg}, \binits{M.}},
\oauthor{\bsnm{Wicke}, \binits{M.}},
\oauthor{\bsnm{Yu}, \binits{Y.}},
\oauthor{\bsnm{Zheng}, \binits{X.}}:
{TensorFlow}: Large-Scale Machine Learning on Heterogeneous Systems.
Software available from tensorflow.org
(2015).
\url{https://www.tensorflow.org/}
\end{botherref}
\endbibitem

\bibitem[\protect\citeauthoryear{Kingsford and Salzberg}{2008}]{kingsford2008decision}
\begin{barticle}
\bauthor{\bsnm{Kingsford}, \binits{C.}},
\bauthor{\bsnm{Salzberg}, \binits{S.L.}}:
\batitle{What are decision trees?}
\bjtitle{Nature biotechnology}
\bvolume{26}(\bissue{9}),
\bfpage{1011}--\blpage{1013}
(\byear{2008})
\end{barticle}
\endbibitem

\bibitem[\protect\citeauthoryear{Rabaey et~al.}{2002}]{rabaey2002digital}
\begin{bbook}
\bauthor{\bsnm{Rabaey}, \binits{J.M.}},
\bauthor{\bsnm{Chandrakasan}, \binits{A.}},
\bauthor{\bsnm{Nikolic}, \binits{B.}}:
\bbtitle{Digital Integrated Circuits}
vol. \bseriesno{2},
(\byear{2002})
\end{bbook}
\endbibitem

\end{thebibliography}

\end{document}